\documentclass[letterpaper, 10 pt, conference]{ieeeconf}  

\IEEEoverridecommandlockouts                              

\usepackage{amsmath,amssymb,amsfonts}
\usepackage{graphicx}
\usepackage{booktabs}
\usepackage{multirow}
\usepackage{url}
\usepackage{hyperref}
\usepackage{tikz}

\usepackage{algorithm}
\usepackage{algpseudocode}
\usepackage{cite}
\usepackage{comment}

\renewcommand{\Re}{\mathbb{R}}
\newcommand{\T}{\mathcal T}

\renewcommand{\S}{\mathcal S}
\newcommand{\Nlf}{\mathcal{N}_{\text{leaf}}}
\newcommand{\tR}{t_{\mathsf{R}}}

\newcommand{\figref}[1]{Fig.~\ref{#1}}

\DeclareMathOperator*{\argmax}{arg\,max}

\title{\LARGE \bf
A Sample-Based Approach for Hierarchical Information-Theoretic Compression of Probabilistic Occupancy Grids
}

\author{Zhenyu Jin$^{1}$ and Daniel T. Larsson$^{2}$
\thanks{$^{1}$Z.~Jin is a PhD Student in the Aerospace and Mechanical Engineering Department at the University of Arizona in Tucson, AZ, USA. {\tt\small zhjin320@arizona.edu}
}%
\thanks{$^{2}$D.~Larsson is an Assistant Professor in the Aerospace and Mechanical Engineering Department at the University of Arizona in Tucson, AZ, USA. {\tt\small dlarsson@arizona.edu}
}%
}

\begin{document}

\maketitle
\thispagestyle{empty}
\pagestyle{empty}

\begin{abstract}

We develop a sample-based framework for constructing information-driven hierarchical multi-resolution representations of probabilistic occupancy grids.
Recent methods compute information-optimal abstractions via dynamic-programming-based exhaustive recursions, which become computationally prohibitive for large-scale grids and are ill-suited to robotics applications.
To address this limitation, we introduce a sample-based strategy inspired by Monte Carlo Tree Search (MCTS) that incrementally constructs hierarchical abstractions through statistical estimation rather than exhaustive enumeration.
The proposed method is anytime in nature, allowing computation to be terminated at any stage to produce a valid compressed representation.
We compare our approach with the information-optimal Q-tree search algorithm and demonstrate its effectiveness in rapidly generating abstractions of large real-world probabilistic occupancy grids.

\end{abstract}

\section{Introduction}

Autonomous systems are increasingly deployed across a diverse set of applications such as search and rescue, warehouse and supply chain logistics, and agriculture~\cite{lou2004study,queralta2020collaborative,bechar2016agricultural}.
Furthermore, recent advances in sensing and onboard computation have enabled increasingly capable autonomous platforms, leading to a diverse ecosystem of systems that vary widely in size, agility, sensing modalities, and computational resources.
As autonomy expands into increasingly diverse domains, there is a growing need for principled frameworks that explicitly account for sensing, communication, and computational resource constraints.
A central component of autonomy is environment representation, or more specifically, how agents process sensory data from cameras and LiDAR, for example, into structured models that support tasks such as perception and collision-free planning.
In this context, a fundamental challenge in building intelligent autonomous systems is enabling agents to distinguish between task-relevant and irrelevant details within the environment.
Effective autonomy under resource constraints therefore requires principled methods for distilling information into task-relevant representations while respecting computational and memory limitations.
In this work, we focus on one such representation, namely probabilistic occupancy grids, and study the problem of systematically compressing them in an information-driven, resource-aware manner.
The compression of grid representations for autonomous systems applications has been studied extensively, as reducing the size of these representations lowers on-board memory requirements and mitigates the computational burden associated with planning and exploration~\cite{hauer2015multi,hauer2016reduced,Nelson2015,CowlagiTsiotras2008Multiresolution,KambhampatiDavis1986Multiresolution,Behnke2004LocalMultiresolution}.
However, until recently, the design of multi-resolution representations for probabilistic occupancy grids has remained largely heuristic, relying on statistical criteria and human judgment to guide pruning or compression procedures~\cite{EinhornSchroterGross2011GridMapping,CowlagiTsiotras2007BeyondQuadtrees,KraetzschmarEtAl2004ProbabilisticQuadtrees}.
While heuristic approaches often provide practical benefits in terms of execution time, they offer no formal guarantees on the size of the resulting compressed representation or on the amount of information about the variable of interest (e.g., occupancy probability) that is preserved after compression.
In contrast, recent work~\cite{larsson2020q,larsson2021information,larsson2023linear} adopts a more systematic approach to abstraction design for probabilistic occupancy grids by leveraging concepts from information theory.
Specifically, the authors of~\cite{larsson2020q} formulate the design of hierarchical multi-resolution representations as an optimization problem inspired by the information bottleneck (IB) principle. The information bottleneck problem, first introduced in~\cite{tishby2000IBmethod}, with additional variants introduced in~\cite{StrouseSchwab2017DeterministicIB,SlonimTishby1999AIB}, is an encoder-design framework that balances two competing objectives: (i) retaining predictive information about a designated relevant variable in the compressed representation, and (ii) achieving a desired level of compression.
The IB formulation is probabilistic, quantifying these objectives using mutual information between the corresponding random variables, assuming access to their joint distribution.
A key observation in~\cite{larsson2020q} is that a hierarchical occupancy grid representation can be interpreted as a deterministic signal encoder with a particular structure. 
This perspective establishes a bridge between information-theoretic signal compression and hierarchical abstraction design, enabling information-theoretic quantities such as mutual information to be defined over structured encoders (e.g., quadtrees and octrees). 
In doing so, it provides a rigorous mechanism for quantifying the information retained by a multi-resolution representation about cell occupancy.
To solve the IB-inspired occupancy grid compression problem, an action-value function reminiscent of those commonly used in reinforcement learning is introduced, where the decision at each node is whether to expand (refine) or prune (compress) the corresponding region.
The resulting algorithm, termed Q-tree search, is shown to return an optimal solution to the IB-based compression problem. However, the method is recursive in nature and requires a bottom-up traversal of the entire hierarchical structure.
As a result, although the approach yields a provably optimal solution to the probabilistic grid-compression problem, it is computationally expensive, particularly for the large-scale grids commonly encountered in real-world robotics applications.
It is therefore that in this paper we develop a method that enables autonomous systems to design information-informed hierarchical multi-resolution abstractions of probabilistic occupancy grids while remaining responsive to the computational constraints of real-world robotics.
To this end, we propose a sample-based strategy inspired by Monte Carlo Tree Search (MCTS) to construct multi-resolution information-driven compressions over large domains.
Our method is anytime in nature, allowing the abstraction process to be queried at any stage to obtain a valid compressed representation.
By adopting a sample-based search procedure, we avoid exhaustive bottom-up recursions over the full hierarchical structure while naturally accommodating heuristics that accelerate the discovery of high-quality solutions.
Consequently, the proposed framework enables a principled trade-off between computational effort and abstraction quality in the design of information-driven multi-resolution occupancy grid representations.
%

\section{Problem Formulation \& Key Challenges}\label{sec:ProblemFormulation_KeyChallenges}

%
As our objective is to design hierarchical multi-resolution representations of probabilistic occupancy grids, we first introduce the probabilistic framework underlying the problem.
Let $(\Omega, \mathcal F, \mathbb P)$ be a probability space, and let $X: \Omega \to \Re$, $Y: \Omega \to \Re$, and $T: \Omega \to \Re$ be random variables with joint distribution $p(x,y,t) = \mathbb{P}(\{\omega\in\Omega: X(\omega)=x,Y(\omega)=y,Z(\omega)=z\})$.
The roles of these variables in our setting are described below.
Our developments will adopt the information bottleneck (IB) principle~\cite{tishby2000IBmethod}, which considers the problem
\begin{equation}\label{eq:IBprob_std}
    \max_{p(t|x)} I(T;Y) - \alpha I(T;X),
\end{equation}
where $X$ denotes the original (source) signal, $T$ the compressed representation of the source $X$, $Y$ the relevance (auxillary) variable for which we wish to maintain predictive information regarding when designing the signal encoder $p(t|x)$.
The IB problem assumes the Markov condition $Y \to X \to T$, meaning that $T$ and $Y$ are conditionally independent given $X$.
This reflects the fact that the representation $T$ is constructed solely from $X$, and so any information $T$ retains about $Y$ must be mediated through $X$.
The encoder $p(t|x)$ maps outcomes of $X$ to those of $T$, potentially stochastically, so as to retain predictive information about $Y$ (measured by $I(T;Y)$) while promoting compression (measured by $I(T;X)$), where the parameter $\alpha>0$ governs the trade-off between information retention and compression, and $I(X;Y)$ is the mutual information between random variables $X$ and $Y$~\cite{CoverThomas2006ElementsIT}.
The objective of the IB problem~\eqref{eq:IBprob_std} is thus to design a compressed representation $T$ by specifying the encoder $p(t|x)$ so as to balance the retainment of predictive information regarding $Y$ (maximizing $I(T;Y)$) while compressing the source $X$ to as high degree as possible (by minimizing $I(T;X)$), where the balance between the two is specified by $\alpha$.
Note that~\eqref{eq:IBprob_std} does not enforce any structural constraints on the encoder $p(t|x)$ required for occupancy grid representations into grid structures amenable to autonomous systems applications, and thus we will restrict the encoder class to hierarchical multi-resolution trees.
Moreover, since our goal is to form hierarchical, multi-resolution, compressions of probabilistic occupancy grids, we next introduce the formalism that relates the information-theoretic concepts above to that of probabilistic grids. 
To this end, assume that, for some integer $n >0$, the environment is represented by a $2^n \times 2^n$ probabilistic grid $\mathcal X \in \Re^{2^n \times 2^n}$, where each finest-resolution cell $(\mathcal X)_i$ has occupancy probability $o_i \in [0,1]$.
We define the source $X$ so that its outcomes correspond to finest-resolution cells $x_i$, and define $Y \in \{0,1\}$ as the occupancy variable with conditional distribution given by $p(y=1|x_i)=o_i$, which is specified by the occupancy map.
Restricting the IB optimization to the space $\mathcal T^{\mathcal S}$ of feasible hierarchical trees (e.g., quadtrees or octrees) by constraining the allowable structure of $p(t|x)$, the problem we consider for the remainder of the paper can be formulated as
\begin{equation}\label{eq:IBtreeSrchProblem}
    \max_{\mathcal T \in \mathcal T^{\mathcal S}} I_Y(\mathcal T) - \alpha I_X(\mathcal T),
\end{equation}
where for $\T_q\in\T^\S$, $I_Y(\mathcal T_q)=I(T_q;Y)$ and $I_X(\mathcal T_q)=I(T_q;X)$, where $T_q$ is the compressed random variable induced by tree $\mathcal T_q$~\cite{larsson2020q}.
The restriction of $p(t|x)$ to the space of hierarchical representations is possible via the observation that each tree $\T_q \in\T^\S$ defines a deterministic mapping from finest-resolution cells (outcomes $x$ of $X$) to leaf nodes (outcomes $t$ of $T_q$), and thus each $\T_q$ corresponds to a deterministic encoder $p_q(t|x)$ with special structure~\cite{larsson2020q}.
The objective of~\eqref{eq:IBtreeSrchProblem} is therefore to identify a hierarchical, multi-resolution representation that maximizes the retention of (probabilistic) occupancy information while compressing the original grid to as high degree as possible.
It is important to note that, since $\mathcal T^{\mathcal S}$ is a countable set, the problem~\eqref{eq:IBtreeSrchProblem} belongs to a class of discrete optimization problems.
%

%
An optimal solution to~\eqref{eq:IBtreeSrchProblem} can be obtained via the Q-tree search algorithm, a dynamic-programming-inspired algorithm that recursively evaluates an action-value function to determine whether each node should be expanded or pruned~\cite{larsson2020q}.
Although Q-tree search returns provably optimal solutions, it exhibits several limitations that restrict its practical deployment in real-world robotic systems:
\begin{enumerate}
    \item\emph{Creation of Full Tree for Evaluation}: The action-value function is computed via bottom-up recursion from leaf nodes to the root, necessitating generation and evaluation of the entire finest-resolution tree and all subsequent nodes.
    Consequently, an abstraction cannot be produced unless all nodes are evaluated, precluding partial or budget-constrained computation.
    \item\emph{Resource Budget Adaptivity}: Robotic platforms operate under limited computational resources.
    Therefore, a practical abstraction method should permit a controllable trade-off between computation and solution quality.
    As a dynamic-programming procedure, Q-tree search provides no natural mechanism for intermediate solutions.
    Instead, the algorithm must run to completion before returning a solution.
    \item\emph{Limited Parallelization Structure}: The strict bottom-up dependency present in Q-tree search that requires the value function of all children to be evaluated before updating a parent, constrains parallel execution and limits opportunities for hardware acceleration, particularly on resource-constrained edge-computing platforms.
\end{enumerate}
To address these limitations, we propose a sample-based framework for solving~\eqref{eq:IBtreeSrchProblem} inspired by the Monte Carlo Tree Search (MCTS) algorithm~\cite{SuttonBarto2018RL}.
Rather than exhaustively evaluating the action-value function over all nodes of the full tree, the method incrementally estimates action values through stochastically sampled rollouts and adaptively concentrates computation in regions of high estimated utility (high predictive information).
Moreover, the algorithm allows agents to limit the number of stochastic rollouts, thereby providing a means to control the computational effort expended in solving~\eqref{eq:IBtreeSrchProblem}, and may be terminated in an anytime fashion to return a valid solution at any intermediate stage of execution.
The next section presents the formalism of the proposed approach, followed by results that demonstrate its ability to rapidly generate multi-resolution information-driven abstractions of large real-world probabilistic grids.
%

\section{A Sample-Based Algorithm for The Design of Hierarchical Abstractions}\label{sec:sampleBasedAlgDesign}

In this section, we present an incremental sample-based algorithm for solving~\eqref{eq:IBtreeSrchProblem} inspired by Monte Carlo Tree Search (MCTS).
A key observation is that the IB objective over $\T^\S$ decomposes additively over local partition refinements, enabling a node-centric action-value formulation in which the value of a state--action pair corresponds to refining a specific region.
Accordingly, solving~\eqref{eq:IBtreeSrchProblem} can be viewed as incrementally selecting which nodes to expand, allowing the problem to be formulated as a sequential decision-making process.

\subsection{Value-Based Formulation of the Tree-Search Problem}\label{subsec:RLprobFormulation}
Given any $\T_k \in \T^\S$ at time $k$, we let the state $s_k = \Nlf(\T_k)$, where $\Nlf(\T_k)$ is the set of leafs nodes of the tree $\T_k$.
When in the state $s_k$ corresponding to the tree $\T_k$, the agent is permitted to select at most one of the leaf nodes of $\T_k$ for expansion (refinement), and thus the action at time $k$ is $a_k = t$ for some $t \in \Nlf(\T_k)$, or $a_k = -1$ if no expansion is selected.
Note that the feasible action space is state-dependent, and is given by $\mathcal A(s_k) = \Nlf(\T_k) \cup \{-1\}$.
When in state $s_k$ and action $a_k$ is selected, the state transitions to new state (tree) $s_{k+1}$ according to the deterministic relation $s_{k+1} = f(s_k,a_k)$, where 
\begin{equation}\label{eq:f_def}
    f(s,a) =  
    \begin{cases}
        (s \setminus a) \cup \mathcal C(a), & \text{ if } a \neq -1, \\
        s, & \text{ otherwise,}
    \end{cases}
\end{equation}
and $\mathcal C(a)$ is set of children of the node $a$ selected for expansion. 
Note that the tree corresponding to $s_{k+1}$ will differ by only a single leaf-node expansion to the tree given by state $s_k$.

Now, let $J_\alpha(\T) = I_Y(\T) - \alpha I_X(\T)$ be the objective function of problem~\eqref{eq:IBtreeSrchProblem}.
We define the one-step reward function of our process according to
\begin{equation}\label{eq:reward_def}
    R_\alpha(s_k,a_k) = J_\alpha(\T_{k+1}) - J_\alpha(\T_k) = \Delta J_\alpha(a_k),
\end{equation}
where $\T_k = \texttt{tree}(s_k)$ and $\T_{k+1} = \texttt{tree}(s_{k+1}) = \texttt{tree}(f(s_{k},a_k))$ are the trees corresponding to states $s_k$ and $s_{k+1}$, respectively.
Note that when the objective $J_\alpha(\T)$ is evaluated between two trees that differ by the expansion of a single leaf node as in~\eqref{eq:reward_def}, the change in cost depends only on the expanded node $a_k$.
Accordingly, the objective difference may be written as $\Delta J_\alpha(a_k) = \Delta I_Y(a_k) - \alpha \Delta I_X(a_k)$.
Furthermore, the incremental information terms are given by $\Delta I_X(a_k) = p(a_k)H(\Pi)$ and $\Delta I_Y(a_k) = p(a_k)\text{JS}_{\Pi}(p(y|t'_1),\ldots,p(y|t'_{\lvert \mathcal C(a_k) \rvert}))$, where $\text{JS}_{\Pi}(\cdot)$ denotes the Jensen--Shannon (JS) divergence~\cite{Lin1991DivergenceShannonEntropy}.
Here $\{t'_1,\ldots,t'_{\lvert \mathcal C(a_k) \rvert}\}$ are the children of the node $a_k$, $\Pi$ is the merger distribution, and $H(\Pi)$ is the Shannon entropy of $\Pi$~\cite{CoverThomas2006ElementsIT,SlonimTishby1999AIB}.
For additional details regarding these incremental quantities, the interested reader is referred to~\cite{larsson2020q}.
The problem~\eqref{eq:IBtreeSrchProblem} can then be written as
\begin{equation}\label{eq:std_RLTreeProb}
    \max_{\{a_0,s_1,a_1,\ldots,s_N,N\}}\quad\sum_{k=1}^{N} R_\alpha(s_k,a_k),
\end{equation}
where the decision horizon $N$ is unknown a priori, and the initial state is $s_0 = \{\tR\}$, with $\tR$ denoting the root (single-node) tree.
We solve~\eqref{eq:std_RLTreeProb} using a sample-based approach inspired by reinforcement learning.
To this end, we define the state--action value function as
\begin{equation}\label{eq:tree_level_Q}
Q^\pi_\alpha(s,a)= \mathbb{E}_{\pi}\Big[\sum_{j\ge 0} R_\alpha(s_{k+j},a_{k+j}) | s = s_k, a=a_k \Big],
\end{equation}
which represents the expected reward-to-come when starting in state $s=s_k$, selecting action $a=a_k$, and following policy $\pi$ thereafter~\cite{SuttonBarto2018RL}.
The optimal state--action value function is $Q^*_\alpha(s,a) = \max_\pi Q^\pi_\alpha(s,a)$, and the optimal value function is $V^*_\alpha(s) = \max_a Q^*_\alpha(s,a)$.
The optimal action to select in state $s$ is therefore $a^* = \argmax_a Q^*_\alpha(s,a)$.
A key challenge in our setting is the cardinality and representation of the state $s_k$.
The state encodes a large number of possible grid compression alternatives (i.e., trees), and multiple sequences of node expansions may lead to the same state, making the mapping between expansion sequences and states non-unique.
However, the structure of the objective in~\eqref{eq:reward_def} allows the problem to be reformulated at the level of nodes, since the cost depends only on the node selected for expansion and the regions defined by any subset of leaf nodes are disjoint.
Consequently, selecting leaf node $a_k$ for expansion in state $s_k$ yields the following reward-to-come
\begin{equation}\label{eq:optActVal}
    Q_\alpha^*(s_k,a_k) = \sum_{t \in s_k} \delta_{a_k,t} \bar y_{\alpha}^*(t),
\end{equation}
where
\begin{equation}
    \bar y_{\alpha}^*(t) = \Delta J_\alpha(t)+\sum_{c\in\mathcal C(t)} \bar{V}_\alpha^{*}(c),
\end{equation}
and
%
\begin{align}
\bar{V}_\alpha^{*}(t) &= \max\{0,\ \Delta J_\alpha(t)+\sum_{c\in\mathcal C(t)} \bar{V}_\alpha^{*}(c)\}, \label{eq:Qstar_recursion}\\
&=\max\{0,\ \bar y_{\alpha}^*(t)\}, \nonumber
\end{align}
%
and $\delta_{a_k,t}$ is the Kronecker delta function.
Equation~\eqref{eq:Qstar_recursion} corresponds to the node-based value function used in Q-tree search~\cite{larsson2020q}, which computes an optimal solution to~\eqref{eq:std_RLTreeProb} via a bottom-up recursion over the finest-resolution tree.
At each node, the recursion evaluates two alternatives: (i) prune the node (future value $0$), or (ii) expand the node, yielding future value $\Delta J_\alpha(t)+\sum_{c\in\mathcal C(t)} \bar V_\alpha^{*}(c)$.
Thus, refinement of $t$ is beneficial only if the net gain of expansion, including optimal continuation through its children, is positive, as otherwise the node is kept as a leaf.
This decision is determined through $\bar V_\alpha^*(t)$, and nodes with $\bar V_\alpha^*(t)>0$ are expanded in a top-down procedure using~\eqref{eq:optActVal} to select subsequent actions (expansions).\footnote{Notice that if $\bar V^*_\alpha(t) > 0$ then $\bar V^*_\alpha(t) = \bar y_{\alpha}^*(t) > 0$.}
The procedure terminates when no expansion yields positive value in~\eqref{eq:optActVal}, at which point action $a_k=-1$ is selected to stop the search.
As discussed in Section~\ref{sec:ProblemFormulation_KeyChallenges}, although this procedure yields an optimal solution to~\eqref{eq:IBtreeSrchProblem}, it suffers from limitations that restrict its applicability in robotics.
We therefore seek a sample-based alternative, presented next.
%

\subsection{Evaluation and Back-propagation of Sample-Based Rollouts}\label{subsec:evalAndBackprop_rollouts}

The goal of our algorithm is to generate an estimate of $\bar V^*_\alpha(t)$ in~\eqref{eq:Qstar_recursion}, denoted $\hat V_{\alpha}(t)$, using a sample-based Monte Carlo approach.
Our method seeks to obtain these estimates rapidly while avoiding evaluation of nodes unlikely to be retained in the compressed tree.
Importantly, relations~\eqref{eq:optActVal} and~\eqref{eq:Qstar_recursion} show that estimating $\bar V_\alpha^*(t)$, which is a function defined over nodes $t$, is sufficient to determine $Q^*_\alpha(s_k,a_k)$, the latter depends on the tree state $s_k$.
Therefore, the algorithm operates by generating samples directly in the node space.
Our algorithm thus proceeds in two phases.
First, a set of rollout trajectories is generated through the tree to enable forward evaluation of the estimated reward-to-come.
Then, a backward induction phase is used to update the estimate $\hat V_\alpha(t)$ for each node $t$ affected by the rollout.
Moreover, we will assume that only a finite number $M$ of rollout samples are available.
Phase one proceeds as follows.
Suppose the algorithm has generated a tree $\T_k \in \T^\S$ corresponding to state $s_k$.
An action $a_k$ is then selected through an exploration procedure using a two-step strategy that combines Upper-Confidence-Bound (UCB) and $\varepsilon$-greedy methods~\cite{SuttonBarto2018RL}.
In the first exploration step, the UCB rule is used to select a leaf node for expansion according to
\begin{equation}\label{eq:explorationStrategy}
    a_k = \argmax_{a}  \left\{ \hat Q_\alpha(s_k,a) + \kappa\sqrt{\frac{\ln N_1(s_k)}{N_2(s_k,a)}} \right\},
\end{equation}
where $\hat Q_\alpha(s_k,a) = \sum_{t \in s_k} \delta_{a,t} \hat y_{\alpha}(t)$, $\hat y_{\alpha}(t)$ is an estimate of $\bar y_{\alpha}^*(t)$ computed in phase two, $N_1(s_k)$ is the number of visits to state $s_k$, $N_2(s_k,a)$ is the number of times action $a$ has been selected in state $s_k$, and $\kappa$ is a constant that balances exploration and exploitation.
Once a suitable leaf node has been identified by~\eqref{eq:explorationStrategy}, the node is expanded and a rollout is initiated in the node space to generate an estimate of~\eqref{eq:Qstar_recursion}.
During the rollout, an $\varepsilon$-greedy exploration rule recursively determines whether a child of an expanded node is selected for further expansion or if the rollout is terminated.
That is, the rollout ends when the $\varepsilon$-greedy selection chooses not to expand any child of an expanded node, thereby terminating the finite node sequence that constitutes the rollout.
It is important to note that the proposed two-step exploration procedure is designed to use the available rollout samples efficiently when estimating $\bar V^*_\alpha(t)$, similar to the guided rollout-selection mechanism in MCTS.
In particular, the tree-state UCB rule~\eqref{eq:explorationStrategy} directs rollouts toward leaf nodes that appear promising while balancing sample allocation with exploring under-explored regions (leafs).
In contrast, purely node-centric rollouts could waste many samples estimating $\bar V_\alpha^*(t)$ for nodes $t$ that provide little information gain or are unlikely to appear in a solution to~\eqref{eq:std_RLTreeProb}.
Thus, combining tree-level (UCB) and node-level ($\varepsilon$-greedy) exploration focuses rollouts on informative regions and supports incremental construction of the hierarchical tree.
Phase two begins once the rollout process described above terminates.
At this point, the algorithm has generated a sequence of nodes $(t_0,\ldots,t_\ell)$, where $\ell$ is the variable length of the rollout sequence.
We then use the one-step reward information, $\Delta J_\alpha(t_j)$, for each of these nodes to update our estimate of the function $\hat V_\alpha(t_0)$, as follows.
For a general rollout sequence $(t_0,\ldots,t_\ell)$ starting in node $t_0$, we compute
\begin{equation}\label{eq:rolloutAve}
   \hat y^i_\alpha(t_0) = \hat y^{i-1}_\alpha(t_0) + \frac{1}{N_3(t_0)} \sum_{j=0}^\ell \Delta J_\alpha(t_j),
\end{equation}
where $N_3(t_0)$ is the number of times the node $t_0$ has been visited as part of any rollout, and $i$ is the iteration number.
We then initialize a backpropagation procedure to update $\hat y_\alpha(t)$ for each ancestor $t$ of $t_0$ according to
\begin{equation}\label{eq:backPropRollout}
     \hat y^i_\alpha(t) = \hat y^{i-1}_\alpha(t) + \frac{1}{N_3(t)}\big[\Delta J_\alpha(t) + \sum_{c\in\mathcal C(t)} \max\{0,~\hat y^i_\alpha(c)\}  \big].
\end{equation}
The estimate $\hat V_\alpha(t)$ of $\bar V^*_\alpha(t)$ at iteration $i$ is then given by
\begin{equation}\label{eq:valueEstUpdate}
    \hat V_{\alpha}(t) = \max\{0,~\hat y^i_\alpha(t)\}.
\end{equation}
Once~\eqref{eq:rolloutAve}-\eqref{eq:backPropRollout} have been computed for $t_0$ and each of its ancestors $t$, the process repeats by returning to phase one to again randomly sample a new leaf node for evaluation via~\eqref{eq:explorationStrategy} using the updated action-value function $\hat Q_\alpha(s_k,a) = \sum_{t \in s_k} \delta_{a_k,t} \hat y_{\alpha}(t)$ with $\hat y_{\alpha}(t) = \hat y^i_{\alpha}(t)$.

\subsection{A Monte-Carlo Algorithm for Abstraction Design}

Our proposed approach is summarized in Algorithm~\ref{alg:mcts_abstraction}.
The algorithm is initialized with the root node $\tR$, the rollout budget $M$, and parameters $\alpha$, $\varepsilon$, and $\kappa$, where $\alpha$ controls the information–compression trade-off in~\eqref{eq:IBtreeSrchProblem} and $\varepsilon$, $\kappa$ regulate exploration.
The method incrementally constructs a Monte-Carlo tree $\T_{\text{MC}}$ that guides sampling toward regions likely to contain high-value information.
The tree $\T_{\text{MC}}$ is initialized with the root tree $\tR\in\T^\S$, corresponding to the initial state $s_0=\{\tR\}$.
New states $s$ are added to $\T_{\text{MC}}$ as they are discovered, and their action-value functions~\eqref{eq:optActVal} estimated via the rollout and backpropagation procedure described in Section~\ref{subsec:evalAndBackprop_rollouts}.
The algorithm proceeds in the two phases described earlier.
Phase~I applies the UCB rule~\eqref{eq:explorationStrategy} to explore actions in previously visited states stored in $\T_{\mathrm{MC}}$, as checked by $\texttt{isIn}(\cdot)$ in line~\ref{alg:checkMCTreeState}.
If an unseen state $s'$ is encountered, it is added to $\T_{\mathrm{MC}}$ in line~\ref{alg:addUnseenToMCTree}, after which a rollout is initiated from the children of the expanded node $a$ in line~\ref{alg:rolloutSeq}.
Phase~II then evaluates the rollout using~\eqref{eq:rolloutAve} (line~\ref{alg:rolloutEval}) and propagates value estimates using~\eqref{eq:backPropRollout} (line~\ref{alg:backprop}).
Once the sample budget $M$ is exhausted, node-value estimates are computed via~\eqref{eq:valueEstUpdate} in line~\ref{alg:valueEst}, and the final grid compression tree $\T$ is obtained in line~\ref{alg:treeConstruct} by expanding nodes $t$ for which $\hat V_\alpha(t) > 0$ in a top-down manner starting at the root $\tR$ until no further positive-value expansions remain.

\begin{algorithm}[t]
\caption{Sample-based information-theoretic multi-resolution grid compression.}
\label{alg:mcts_abstraction}
\begin{algorithmic}[1]
\Require root node $\tR$, rollout budget $M$, constants $\alpha,\epsilon,\kappa$.
\Ensure Hierarchical Tree $\mathcal{T} \in \T^\S$.

\State Initialize $\mathcal{T}_{\text{MC}}$ with root $\{\tR\}$.
\State Set $N_1(s)=0$, $N_2(s,a)=0$, and $\hat y_\alpha^0(\tR)=0$.

\For{$i=1$ to $M$}
    \State $s \gets \texttt{getRoot}(\mathcal{T}_{\text{MC}})$;

    \While{true}
        \State $\mathcal{A}(s) \gets \texttt{setActionSpace}(\mathcal{N}_{\mathrm{leaf}}(\texttt{tree}(s)))$;
        \State $\hat Q_\alpha(s,a) \gets \texttt{uptActVal}\!\left(\mathcal{A}(s),\{\hat y_\alpha(t)\}_{t\in s}\right)$;
        \State $a \gets \texttt{lfSel}\!\left(\hat Q_\alpha(s,a),\kappa,N_1(s),N_2(s,a)\right)$;

        \If{$\texttt{isIn}(f(s,a),\mathcal{T}_{\text{MC}})$} \label{alg:checkMCTreeState}
            \State $s \gets f(s,a)$;
        \Else
            \State $s' \gets f(s,a)$;
            \State $\T_{\text{MC}} \leftarrow \texttt{add}(s',\mathcal{T}_{\text{MC}})$;\label{alg:addUnseenToMCTree}
            \State $(t_0,\ldots,t_\ell) \gets \texttt{rollout}(\varepsilon,\mathcal{C}(a))$;\label{alg:rolloutSeq}
            \State $\hat y_\alpha^i(t_0) \gets \texttt{rolloutEval}(t_0,\ldots,t_\ell)$;\label{alg:rolloutEval}
            \State $\hat y_{\alpha}^i(t) \leftarrow \texttt{backProp}(\texttt{ancestor}(t_0))$;\label{alg:backprop}
            \State \textbf{break}
        \EndIf
    \EndWhile
\EndFor
\State $\hat V_\alpha \leftarrow \texttt{createValueEstimate}(\hat y^{M}_\alpha)$; \label{alg:valueEst}
\State \Return $\mathcal{T}= \texttt{createAbstraction}(\hat V_\alpha,\tR)$; \label{alg:treeConstruct}
\end{algorithmic}
\end{algorithm}

\section{Results and Discussion}\label{sec:evaluationAndResults}

In this section, we evaluate the proposed sample-based approach for constructing hierarchical representations of probabilistic grids in two scenarios.
First, we compare our method with the optimal Q-tree search algorithm on a sample grid and show that the resulting abstractions closely match those obtained by exhaustive Q-tree search.
Second, we demonstrate the ability of our approach to generate regional multi-resolution compressions of probabilistic grids constructed from real-world LiDAR data.
%

\subsection{Comparison with Q-tree Search}

\begin{figure}[t]
    \centering
    \includegraphics[width=\columnwidth]{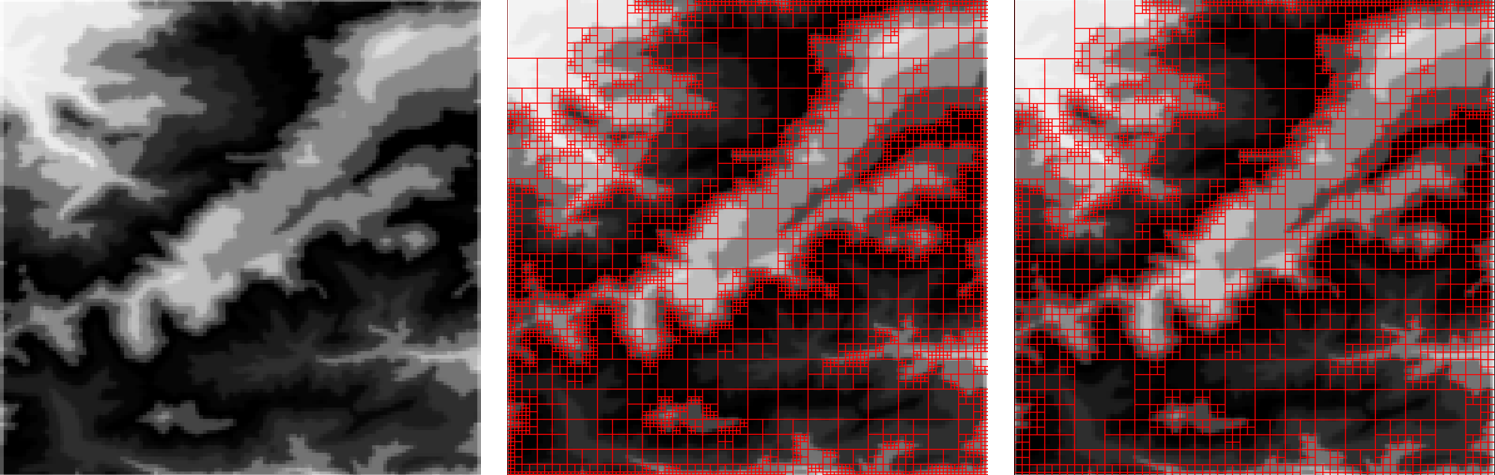}
    \caption{(left) $128 \times 128$ example environment for comparison with Q-tree search. (middle) Abstraction obtained with Q-tree search. (Right) Abstraction obtained with sample-based method with $M = 3000$ rollout budget. Abstractions obtained for $\alpha = 0.0067$.}
    \label{fig:Sample_env_qtree_sampleBased}
\end{figure}

\begin{figure}[t]
    \centering
    \includegraphics[width=0.95\columnwidth]{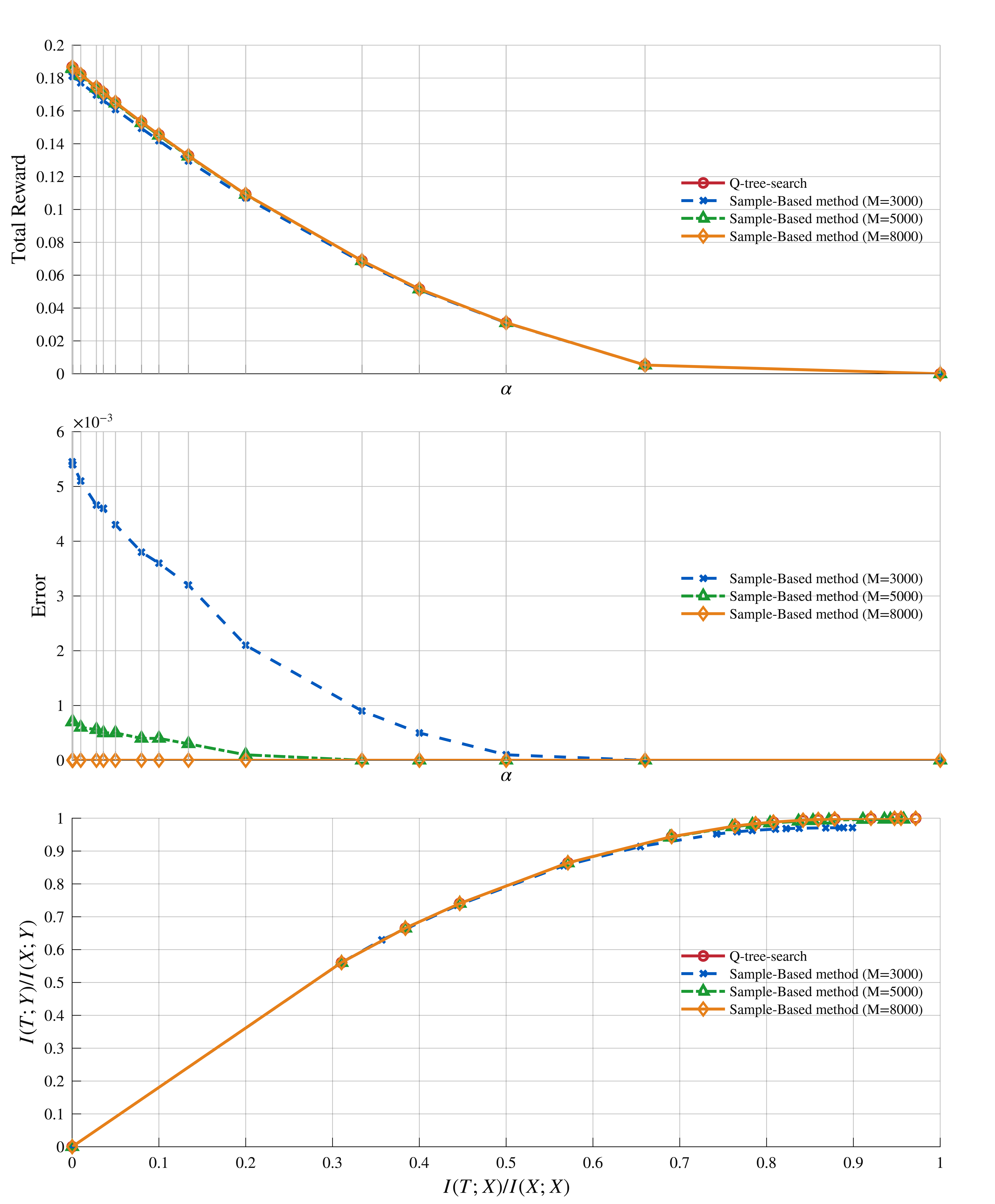}
    \caption{(top) Total reward obtained by executing Q-tree search and the sample-based approach for a variety of rollout budgets $M$ as a function of $\alpha$. (middle) The difference in objective function value (error) between the optimal Q-tree search solution and our sample-based approach as a function of $\alpha$.
    (bottom) the normalized information-plane showing the trade-off between compression (x-axis) and information-retention (y-axis) for each method. The curve is traced by varying $\alpha$, where movement to the right (decreasing $\alpha$) corresponds to less compressed trees (more leaves) that retain more information.}
    \label{fig:fig1}
\end{figure}

\begin{figure}[t]
    \centering
    \includegraphics[width=\columnwidth]{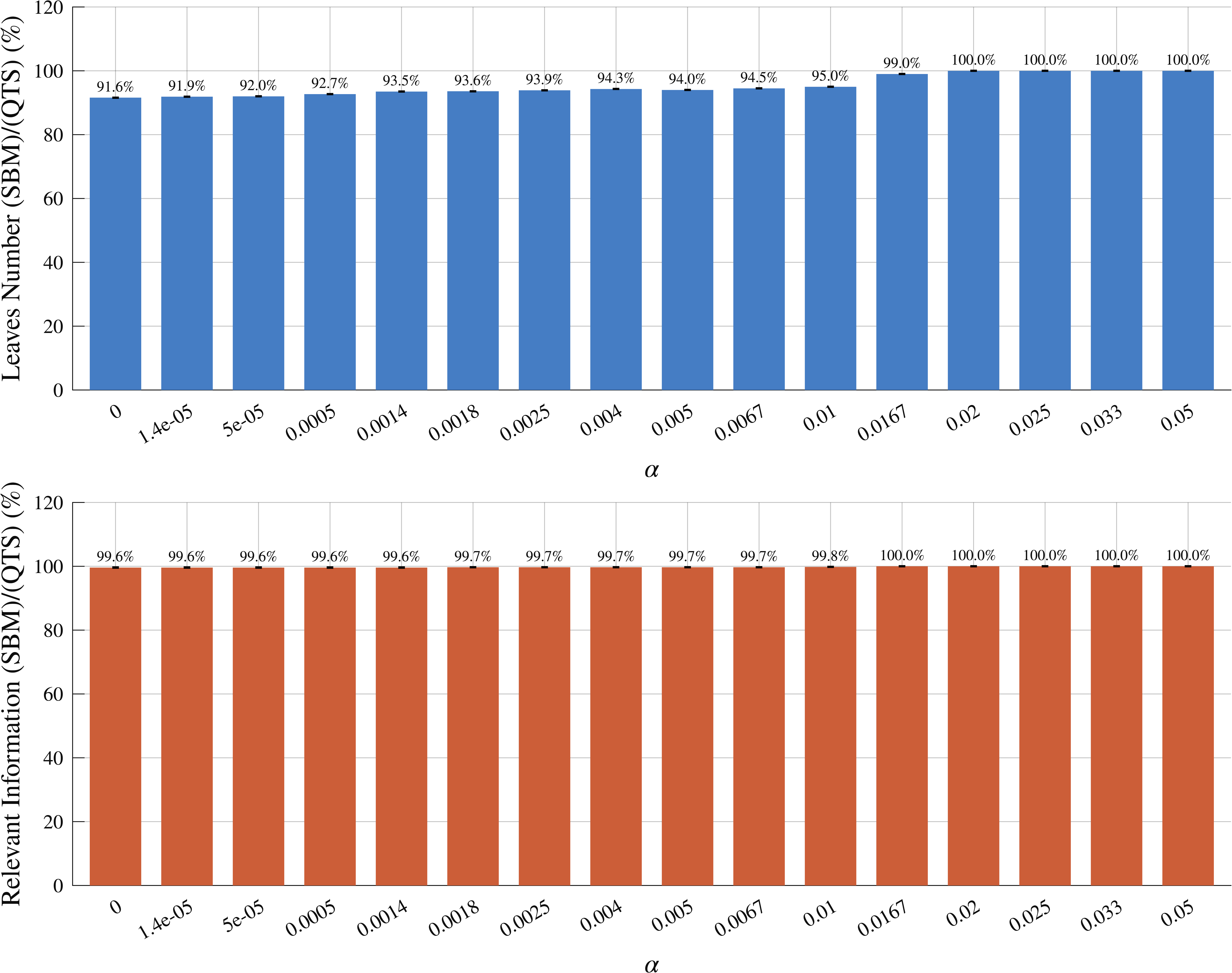}
    \caption{Comparison of the percentage of leaf nodes (top) and relevant information retention $I(T;Y)$ (bottom) between our sample-based method (SBM) and Q-tree search (QTS) for various values of $\alpha$ with $M = 5000$.}
    \label{fig:fig2}
\end{figure}

In this section, we compare our sample-based approach with the Q-tree search method developed in~\cite{larsson2020q}.
We consider the $128\times128$ grid shown in Fig.~\ref{fig:Sample_env_qtree_sampleBased} (left), where grayscale intensity indicates the probability that cell $i$ is occupied, $o_i = p(y=1|x_i)$, with lighter regions corresponding to higher occupancy probability.
The source distribution $p(x)$ is assumed uniform, and so $p(x,y)=p(y|x)p(x)$.
Algorithm~\ref{alg:mcts_abstraction} is executed using $\varepsilon=0.2$ and $\kappa=1$, while the parameter $\alpha$ is varied across several cases.
%
In~\figref{fig:fig1} we report the total reward~\eqref{eq:IBtreeSrchProblem}, the error relative to Q-tree search, and the normalized $I(T;Y)$ versus $I(T;X)$ (information-plane) curves for a range of $\alpha$ and rollout budgets $M$.
The results show that even with a limited rollout budget our method achieves only small error relative to the optimal Q-tree reward.
As the rollout budget increases, the solutions produced by the proposed approach become nearly identical to those of Q-tree search, illustrating the controllable trade-off between computational effort and optimality.
In addition, the information-plane curves recover the same information--complexity frontier as the optimal quadtree solution when sufficient rollout samples are available.
In~\figref{fig:fig2} we compare the number of leaf nodes and the retained relevant information for solutions produced by our sample-based method and Q-tree search across several values of $\alpha$ under a rollout budget of $M=5000$.
The results show that even with a limited rollout budget, the proposed method produces quadtree structures with leaf counts highly consistent with those of the optimal Q-tree solutions.
Moreover, the close agreement in the corresponding values of $I(T;Y)$ indicates that the resulting abstractions retain nearly identical predictive information, demonstrating the similarity between the multi-resolution representations produced by both methods.

\subsection{Application to Real-World Occupancy Grid Compression}

\begin{figure}[b]
    \centering
    \includegraphics[width=\columnwidth]{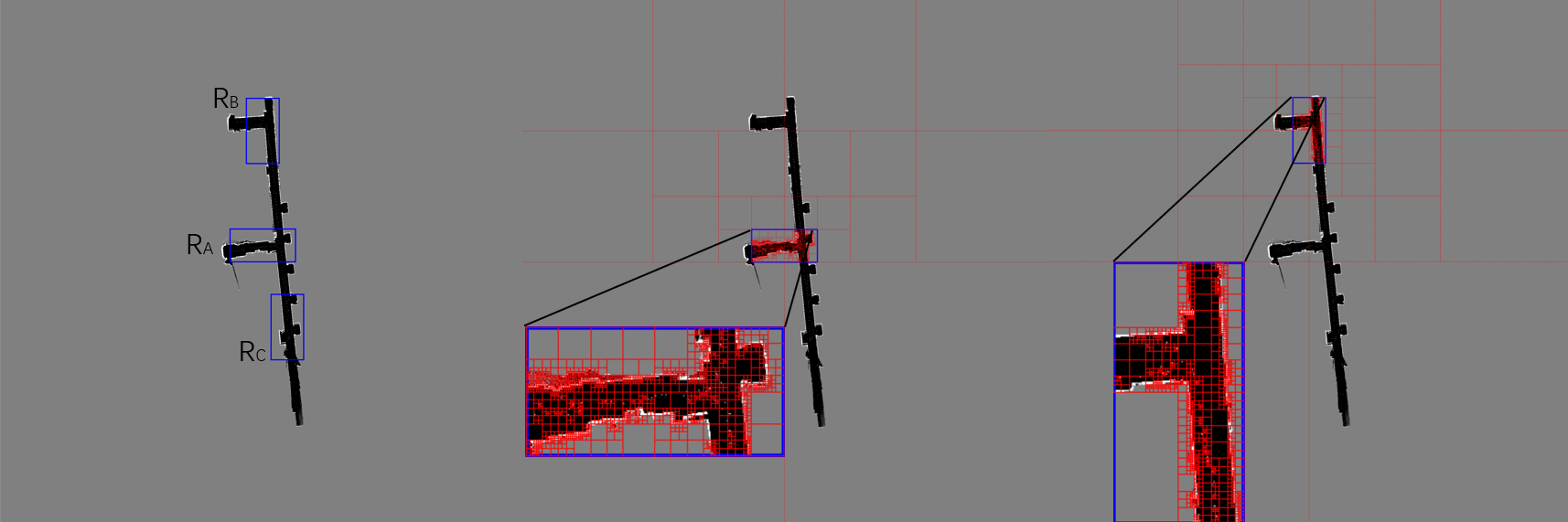}
    \caption{(left) Probabilistic occupancy grid constructed from real-world LiDAR data with three regions of interest labeled as $R_A$, $R_B$, and $R_C$ shown. (middle) and (right) show example multi-resolution compressions of $R_A$ and $R_B$, respectively, obtained by our sample based approach.}
    \label{fig:fig3}
\end{figure}

In this section, we demonstrate the ability of the proposed method to generate multi-resolution compressions of probabilistic occupancy maps constructed from real-world data.
The map shown in \figref{fig:fig3}(left) was generated by an autonomous ground robot navigating the hallway outside our laboratory using ROS2 Foxy and SLAM Toolbox, and has resolution $2048\times2048$.
From this map we select three representative regions of interest (ROIs), denoted $R_A$, $R_B$, and $R_C$, that are relevant to robotic navigation tasks and are shown in \figref{fig:fig3}(left).
Furthermore, we incorporate a heuristic early-stopping strategy during rollouts to accelerate the sample-based search.
Under this heuristic, a rollout may terminate early if it samples a sequence of nodes forming a subtree with positive value, indicating that expanding the subtree root node (start of rollout) yields positive reward improvement.

\begin{figure}[t]
    \centering
    \includegraphics[width=1\columnwidth]{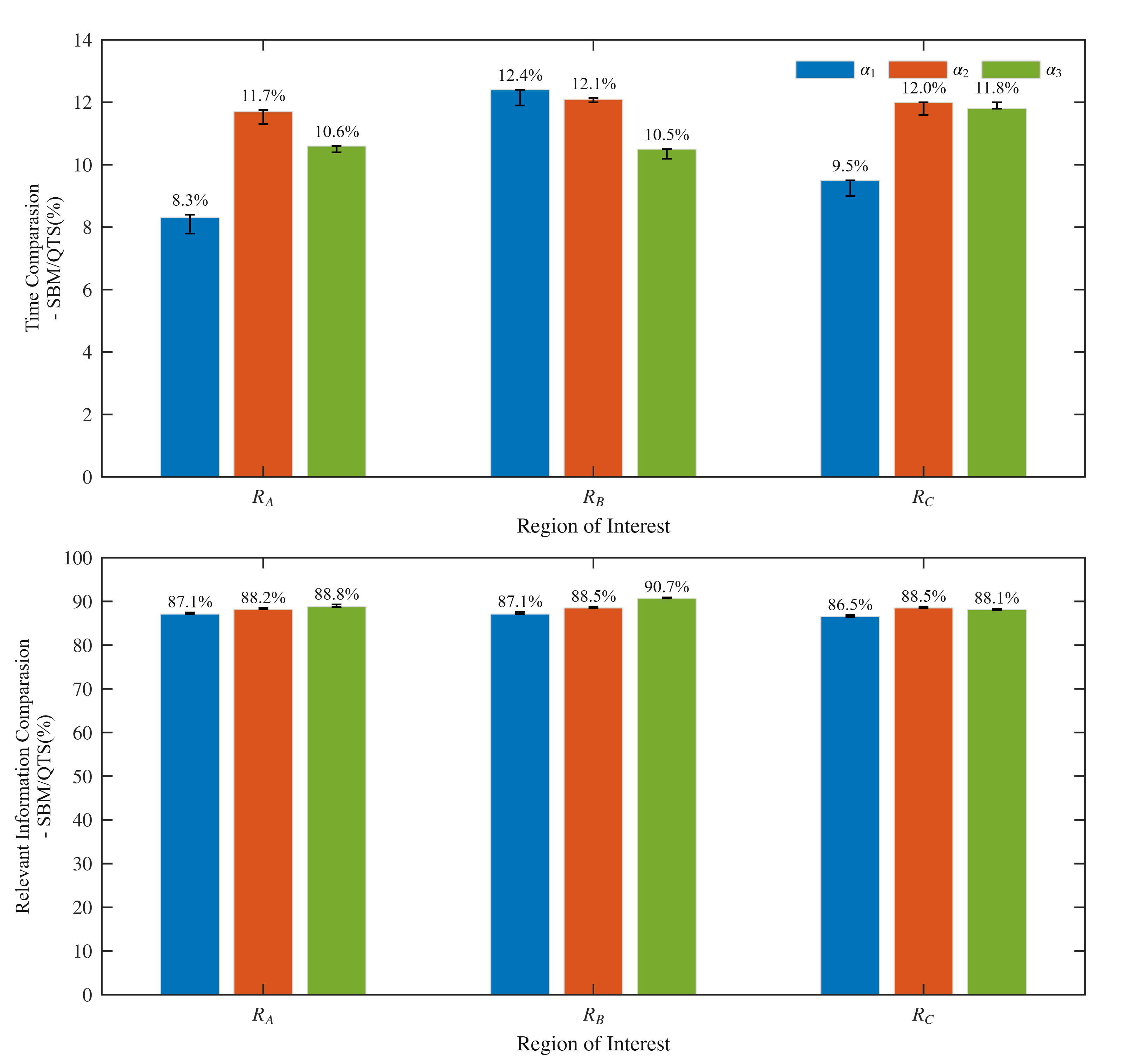}
    \caption{Execution time and relevant information retention comparison between Q-tree search (QTS) and our proposed sample-based method (SBM) for $\alpha_1=0.0005$, $\alpha_2=0.004$, $\alpha_3=0.0067$ and $M = 5500$ over the three regions $R_A$, $R_B$ and $R_C$.}
    \label{fig:fig4}
\end{figure}

In \figref{fig:fig4} we compare the performance of the proposed sample-based method with Q-tree search for generating multi-resolution representations of the regions of interest.
The top panel reports the runtime required to complete the compression task, while the bottom panel shows the amount of task-relevant information preserved after compression.
The results indicate that the sample-based method retains approximately 88\% of the information preserved by Q-tree search while requiring only about 11\% of the corresponding computation time.
In other words, by tolerating a roughly a 12\% reduction in preserved information, the proposed sample-based method achieves an approximate nine-fold speedup in search time compared to Q-tree search.
Notably, these improvements arise entirely from algorithmic design, as no hardware acceleration was used in this experiment.
It is also worth noting that Q-tree search cannot produce a valid solution if the algorithm is interrupted before completion.
In contrast, the proposed method can be terminated at any point during the $M$ rollouts to produce a valid solution, albeit potentially producing a multi-resolution solution with a reduced amount of retained information.
Overall, the results highlight a key advantage of the proposed approach: it enables a flexible trade-off between information optimality and computational effort, allowing substantial runtime improvements with only modest loss in retained information.

\section{Conclusion}\label{sec:conclusion}

In this paper, we developed a sample-based approach for constructing multi-resolution, information-theoretic compression of probabilistic occupancy grids.
Our method leverages ideas from Monte Carlo Tree Search (MCTS) to incrementally build an information-informed hierarchical tree using sampled value estimates.
The proposed approach is anytime in nature, allowing computation to be terminated early while directing the available sampling budget toward regions of the environment that appear most promising.
We demonstrate the effectiveness of our framework through comparison with the information-optimal Q-tree search algorithm and show its ability to rapidly generate abstractions of large-scale probabilistic occupancy grids derived from real-world data.

\bibliographystyle{IEEEtran}


\end{document}